\documentclass[11pt,a4paper]{article}
\usepackage[hyperref]{acl2021}
\usepackage{times}
\usepackage{latexsym}

\usepackage{graphicx}
\usepackage{microtype}
\usepackage{multirow}
\usepackage{booktabs}

\usepackage{hyperref} 
\hypersetup{linkcolor = black}

\aclfinalcopy 

\title{TEXTOIR: An Integrated and Visualized Platform for\\ Text Open Intent Recognition}
\author{
  Hanlei Zhang\textsuperscript{\rm 1}\thanks{\quad  These authors contributed equally to this work.}, Xiaoteng Li\textsuperscript{\rm 1, 2}\footnotemark[1], Hua Xu\textsuperscript{\rm 1}\footnotemark[1] ~\thanks{\quad  Hua Xu is the corresponding author.}, Panpan Zhang\textsuperscript{\rm 1, 2}, Kang Zhao\textsuperscript{\rm 1, 2}, Kai Gao\textsuperscript{\rm 2}\\
  \textsuperscript{\rm 1}State Key Laboratory of Intelligent Technology and Systems, \\ 
Department of Computer Science and Technology, Tsinghua University,\\
  \textsuperscript{\rm 2}School of Information Science and Engineering, Hebei University of Science and Technology\\
    \texttt{zhang-hl20@mails.tsinghua.edu.cn, xuhua@tsinghua.edu.cn}\\
}

\begin{document}
\maketitle
\begin{abstract}
TEXTOIR is the first integrated and visualized platform for text open intent recognition. It is composed of two main modules: open intent detection and open intent discovery. Each module integrates most of the state-of-the-art algorithms and benchmark intent datasets. It also contains an overall framework connecting the two modules in a pipeline scheme. In addition, this platform has visualized tools for data and model management, training, evaluation and analysis of the performance from different aspects. TEXTOIR  provides useful toolkits  and convenient visualized interfaces for each sub-module\footnote{Toolkit code: https://github.com/thuiar/TEXTOIR}, and designs a framework to implement a complete process to both identify known intents and discover open intents\footnote{Demo code: https://github.com/thuiar/TEXTOIR-DEMO}.
\end{abstract}

\section{Introduction}
\begin{figure}[t!]
  \centering  
  \includegraphics[width=0.95\columnwidth ]{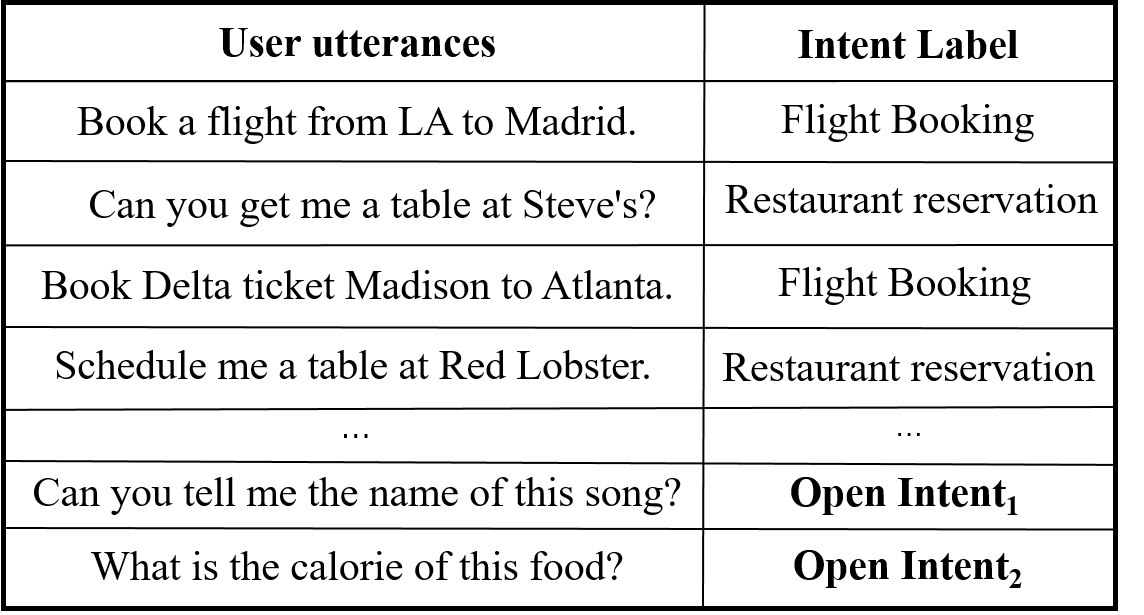}
  \caption{\label{example} An example for Open Intent Recognition.}
\end{figure}

Analyzing user intents plays a critical role in human-machine interaction services (e.g., dialogue systems). However, many current dialogue systems are confined to recognizing user intents in closed-world scenarios, and they are limited to handle the uncertain open intents. As shown in figure~\ref{example}, it is easy to identify specific purposes, such as Flight Booking and Restaurant Reservation. Nevertheless, as the user intents are varied and uncertain, predefined categories may be insufficient to cover all user needs. That is, there may exist some unrelated user utterances with open intents. It is valuable to distinguish these open intents from known intents, which is helpful to improve service qualities, and further discover fine-grained classes for mining potential user needs. 

We divide open intent recognition (OIR) into two modules: open intent detection and open intent discovery. The first module aims to identify n-class known intents and detect one-class open intent~\cite{yan-etal-2020-unknown,lin-xu-2019-deep,Shu2017DOCDO}. It can identify known classes but fail to discover specific open classes. The second module further groups the one-class open intent into multiple fine-grained intent-wise clusters~\cite{10.1145/3366423.3380268,lin2020discovering,perkins-yang-2019-dialog}. Nevertheless, the adopted clustering techniques are not able to identify known categories. 

\begin{figure*}
	\centering
	\includegraphics[scale=.478]{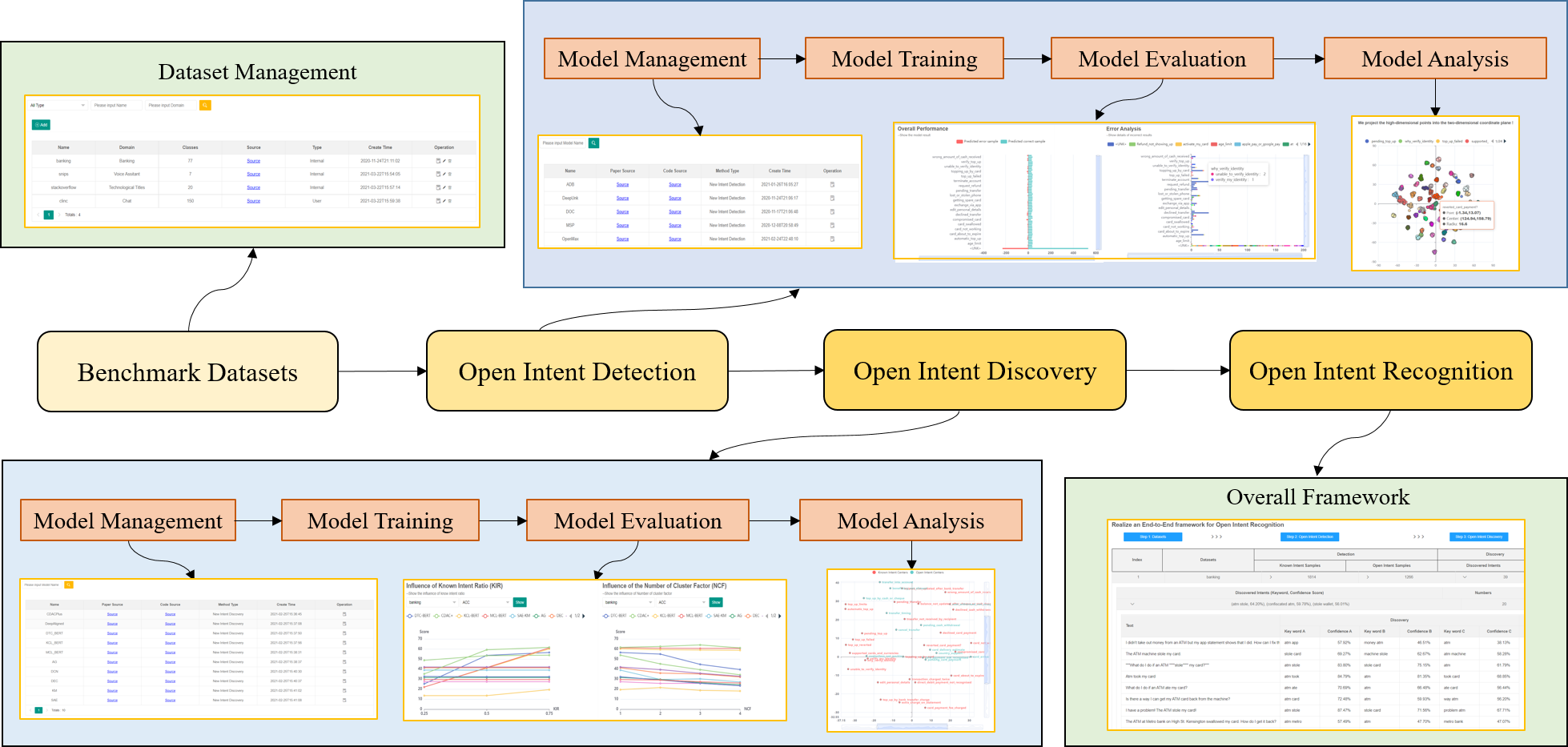}
	\caption{The architecture of the TEXTOIR platform.}
	\label{demo}
\end{figure*}

The two modules have achieved huge progress with various advanced methods on benchmark datasets. However, there still exist some issues, which bring difficulties for future research. Firstly, there are no unified and extensible interfaces to integrate various algorithms for two modules, bringing challenges for further model development. Secondly, the current methods of the two modules lack convenient visualized tools for model management, training, evaluation and result analysis. Thirdly, the two modules both have some limitations for OIR. That is, neither of them can identify known intents and discover open intents simultaneously. Therefore, OIR remains at the theoretical level, and it needs an overall framework to connect the two modules for finishing the whole process.

To address these issues, we propose TEXTOIR, the first integrated and visualized text open intent recognition platform. The platform has the following features:

(1) It provides toolkits for open intent detection and open intent discovery, respectively. The toolkits contain flexible interfaces for data, configuration, backbone and method integration. Specifically, it integrates a series of advanced models for two modules. Each module supports a complete workflow,  including data and backbone preparation with different assigned parameters, training, and evaluation. It provides standard and convenient modules to add new methods. More detailed information can be found on \url{https://github.com/thuiar/TEXTOIR}. 

(2) It designs an overall framework combining two sub-modules naturally, achieving a complete OIR process. The overall framework integrates the advantages of two modules, which can automatically identify known intents and discover open intent clusters with recommended keywords. 

(3) It provides a visualized surface for utilization. Users can leverage the provided methods or add their datasets and models for open intent recognition. We provide the front end interface for the two modules and the pipeline module. Each of the two modules supports model training, evaluation and detailed result analysis of different methods. The pipeline module leverages both the two modules and shows the complete text OIR results. More detailed information can be found on \url{https://github.com/thuiar/TEXTOIR-DEMO}. 

\begin{figure*}
	\centering
	\includegraphics[scale=.478]{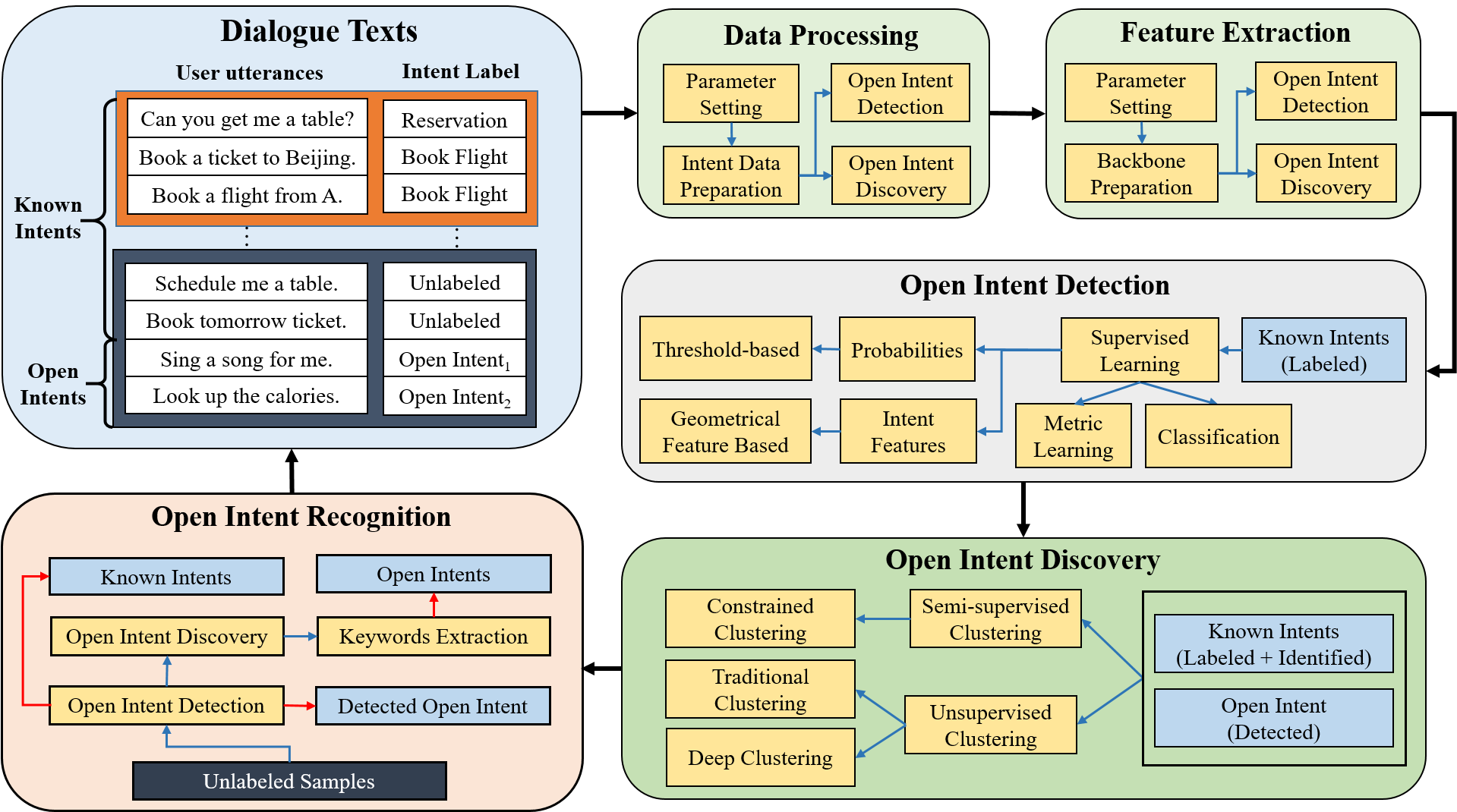}
	\caption{The architecture of Open Intent Recognition.}
	\label{model}
\end{figure*}

\section{Open Intent Recognition Platform}
Figure~\ref{demo} shows the architecture of the proposed TEXTOIR  platform, which contains four main modules. The first module integrates a series of standard benchmark datasets. The 
second and third modules have toolkits for 
both open intent detection and open intent discovery. Besides, it visualizes the whole process (including model management, training, evaluation and result analysis) of two modules. The last module leverages the two modules in a pipeline framework to finish open intent recognition.

\subsection{Data Management}
\label{data}
Our platform supports standard benchmark  datasets for intent recognition, including CLINC~\cite{larson-etal-2019-evaluation}, BANKING~\cite{Casanueva2020},
SNIPS~\cite{DBLP:journals/corr/abs-1805-10190}, and  StackOverflow~\cite{xu-etal-2015-short}. They are all split into training, evaluation and test sets.

 As shown in Figure~\ref{model}, we provide unified data-processing interfaces. It 
 supports preparing data in the format of two modules. For example, it samples known intents and labeled data with the assigned parameters for training and evaluation.  Besides these labeled data, the remaining unlabeled data are also leveraged for open intent discovery. Users can see detailed statistics information from the front-end webpage and manage their datasets.
 
\subsection{Models}
Our platform integrates a series of advanced and competitive models  for two modules, and provides toolkits with standard and flexible interfaces.

\subsubsection{Open Intent Detection}
This module leverages partial labeled known intent data for training. It aims to identify known intents and detect samples that do not belong to known intents. These detected samples are grouped into a single open intent class during testing. We divide the integrated methods into two categories: threshold-based and geometrical feature-based methods. 

The threshold-based methods consist of MSP~\cite{hendrycks17baseline}, DOC~\cite{Shu2017DOCDO}, and OpenMax~\cite{bendale2016towards}. These methods are first pre-trained under the supervision of the known intent classification task. Then, they leverage the probability threshold for detecting the low-confidence open intent samples. The geometrical feature-based methods include DeepUnk~\cite{lin-xu-2019-deep} and ADB~\citep{Zhang_Xu_Lin_2021}. DeepUnk adopts the metric-learning method to learn discriminative intent features, and the density-based methods to detect the open intent samples as anomalies. ADB further uses the boundary loss to learn adaptive decision boundaries. 

\subsubsection{Open Intent Discovery}
This module uses both known and open intent samples as inputs, and aims to obtain intent-wise clusters by learning from similarity properties with clustering technologies. As suggested in~\cite{Zhang_Xu_Lin_Lyu_2021,lin2020discovering}, the integrated methods are divided into two parts, including unsupervised and semi-supervised methods. 

The unsupervised methods include K-Means (KM)~\cite{macqueen1967some}, agglomerative clustering (AG)~\cite{gowda1978agglomerative}, SAE-KM, DEC~\cite{xie2016unsupervised}, and DCN~\cite{yang2017towards}. The first two methods adopt the  Glove~\cite{pennington2014glove} embedding, and the last three methods leverage stacked auto-encoder to extract representations. These methods do not need any labeled data as prior knowledge and learn structured semantic-similar knowledge from unlabeled data. 

The semi-supervised methods include KCL~\cite{hsu2018learning}, MCL~\cite{hsu2018multiclass}, DTC~\cite{Han2019learning},  CDAC+~\cite{lin2020discovering} and DeepAligned~\cite{Zhang_Xu_Lin_Lyu_2021}. These methods can further leverage labeled known intent data for discovering fine-grained open intents. 
\begin{figure*}
    \centering
    \includegraphics[scale=.42]{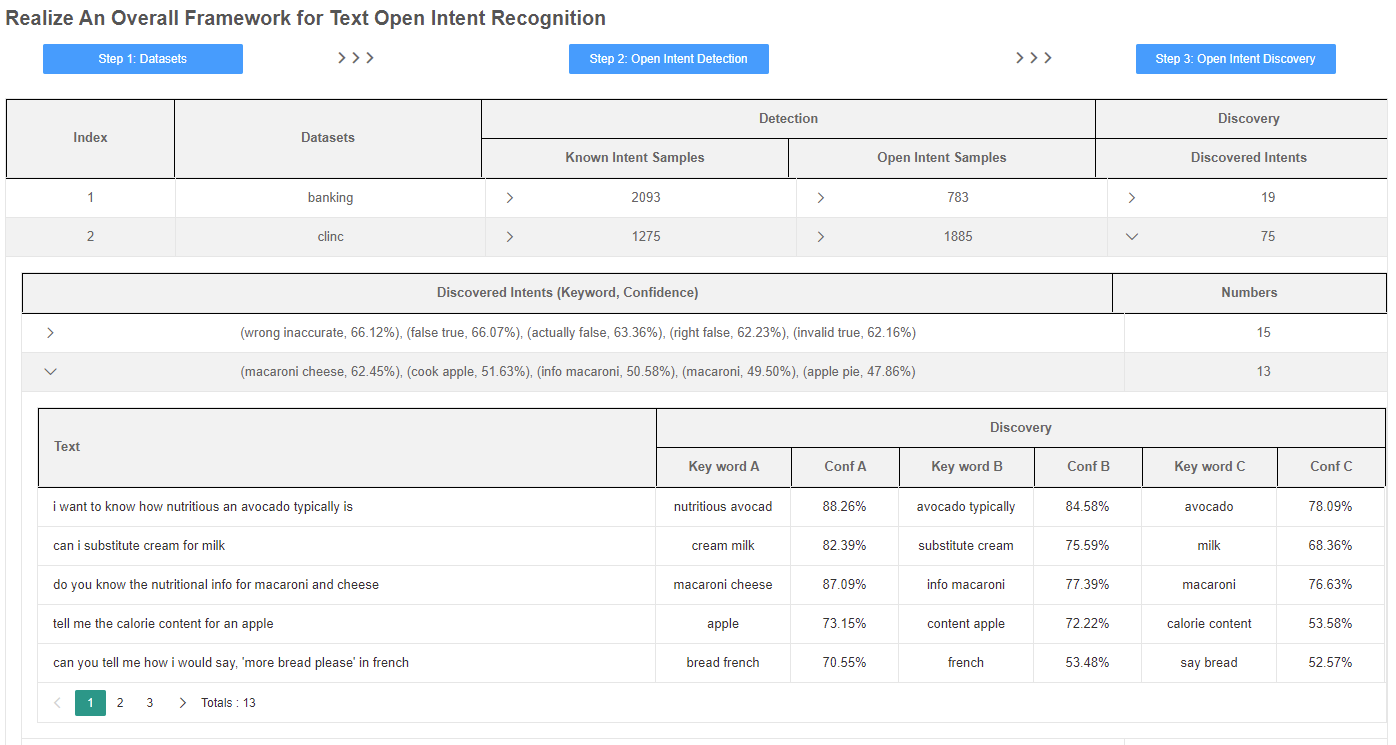}
    \caption{The pipeline framework of open intent recognition.}
    \label{platform}
\end{figure*}

\subsubsection{Interfaces}
We provide a series of interfaces for the two modules. Firstly, the backbones are flexible and unified. For example, the primary backbone is the pre-trained BERT~\cite{devlin2018bert} model, and it supports adding new bert-based models with different downstream tasks. The open intent discovery module also supports other backbones for unsupervised clustering. Secondly, each module has the common data-loaders following the needed formats of the adopted backbones. They encode unified data vectors from the prepared data as mentioned in section~\ref{data}. Thirdly, the parameter configurations are convenient. We extract common parameters (e.g., known intent ratio, dataset, etc.) for each module and support adding different sets of hyper-parameters for  tuning each method. Finally, each approach integrates standard components of training, evaluation, and other specific functions.

\section{Pipeline Framework}
The two modules of open intent detection and discovery are closely related. However, there lacks an overall framework to successively invoke the two modules for both identifying known intents and discovering open intents. TEXTOIR addresses this issue with a proposed pipeline framework, as shown in Figure~\ref{model} and Figure~\ref{platform}. 

The pipeline framework first processes the original data for two modules. Then, it feeds the labeled known intent data to the open intent detection module and trains the selected model by the users. As there is still a mass of unlabeled data containing both known and open intents, it leverages the well-trained open intent detection model to predict the unlabeled training data. The evaluated results on training data contain identified known intents and the detected open intent. We use the predicted known intent data, detected open intent and original labeled data as the inputs of the open intent discovery module.  In this case, the discovery module benefits from the detection module to obtain the augmented inputs for training. Next, the preferred clustering method selected by the users is trained to obtain the open intent clusters. 

After training the two modules, they are used to perform open intent recognition on unlabeled data. Specifically, the well-trained open intent detection method is first used to predict the identified known intents and detected open intent. Then, the open intent discovery method is utilized to predict the detected open intent data to obtain the fine-grained open intent clusters. Finally, the KeyBERT toolkit (mentioned in section~\ref{key}) is leveraged to extract keywords for each open intent cluster with similar-intent sentences. Therefore, our framework identifies known intents and discovers open intent samples in group with keywords as recommended labels.

\section{Visualization}
\subsection{Training and Evaluation}
Our platform provides visualized surfaces for model training and evaluation.  
\begin{figure}[t]
	\centering  
	\includegraphics[scale=.35]{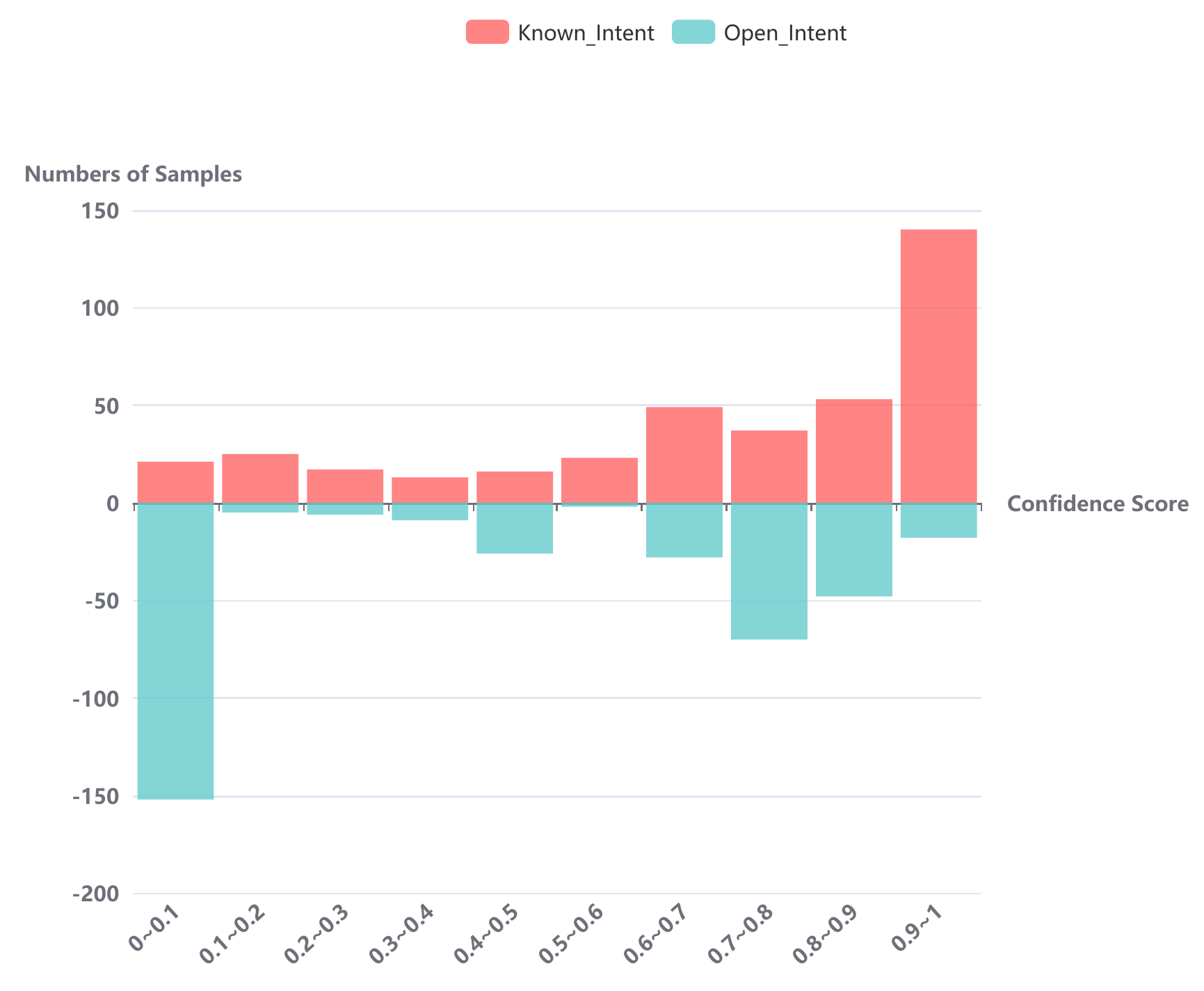}
	\caption{\label{figure-4} Known and Open intent distributions with different confidence scores.}
\end{figure}
For each method, users can change the main hyper-parameters to tune the model.  When training starts, it automatically creates a record for the training  process, which state can be monitored by the users. When the training process finishes successfully, the trained model and related parameters are saved for further utilization. 

For model evaluation, the predicted results are observed from different views. Firstly, the overall performance is shown with the number of correct and false samples for each intent class. On this basis, the number of fine-grained false-predicted classes is further shown to analyze the easily-confused intents regarding the ground truth. Secondly, the influences of the known intent ratio and labeled ratio are correspondingly shown with line charts. Users can observe the results on different selected datasets and evaluation metrics.
\begin{figure}[t]
	\centering  
	\includegraphics[scale=.33]{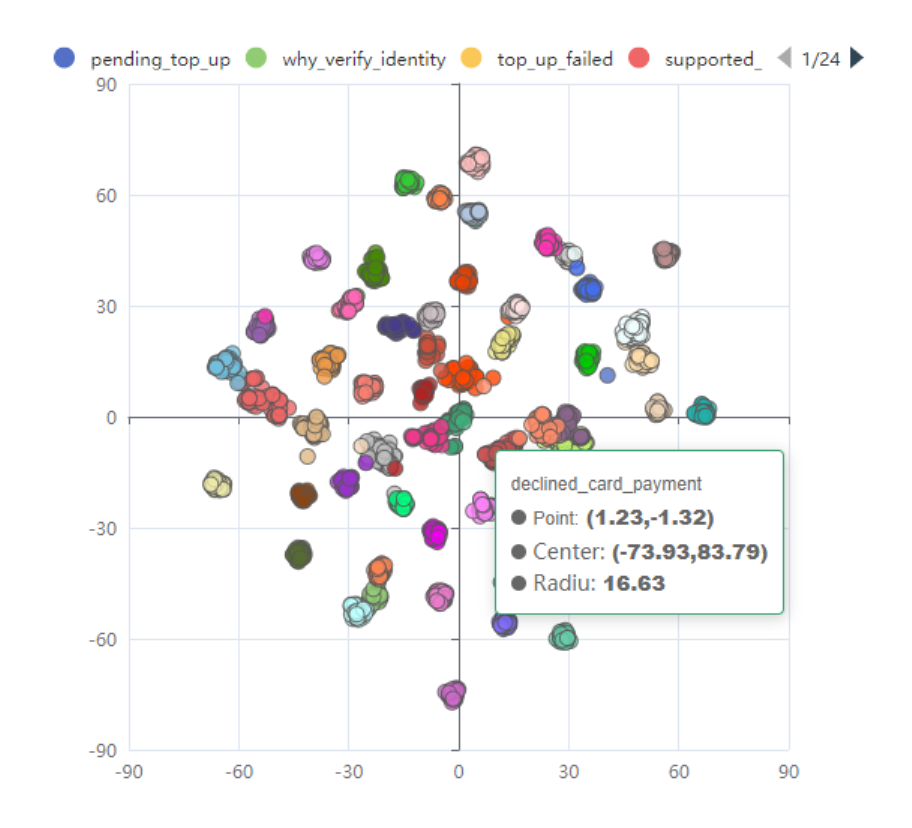}
	\caption{\label{figure-5} Visualization of the intent representations.}
\end{figure}
\begin{figure}
    \centering
    \includegraphics[scale=.40]{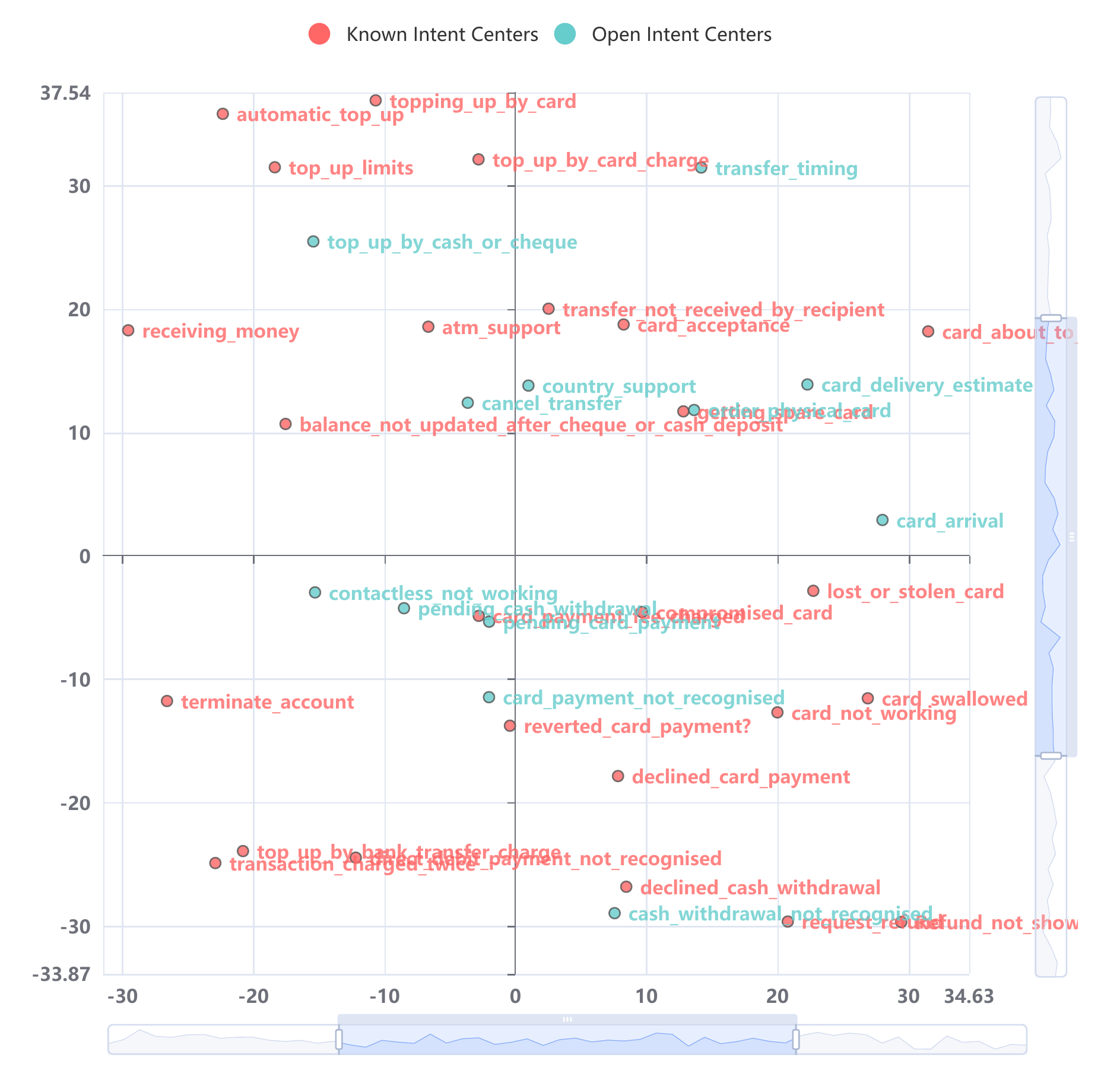}
    \caption{Intent center distribution.}
    \label{figure-6}
\end{figure}
\subsection{Result Analysis}
\begin{table*}[t!]\small
\centering
\begin{tabular}{@{\extracolsep{4pt}}cllccccccccc}
\toprule
\centering
  &  \multicolumn{2}{c}{ADB + DeepAligned} & \multicolumn{2}{c}{CLINC} & \multicolumn{2}{c}{BANKING} & \multicolumn{2}{c}{SNIPS} &\multicolumn{2}{c}{StackOverflow}\\
 \addlinespace[0.1cm] \cline{2-3} \cline{4-5} \cline{6-7} \cline{8-9} \cline{10-11} \addlinespace[0.1cm]
 & KIR  & LR & Known  & Open  & Known & Open & Known & Open & Known & Open\\
\midrule
& 25\% & 50\%  & 89.65 & 86.53 & 84.61 & 63.50 & 87.68 & 32.05 & 82.60 & 45.48 \\
& 25\% & 100\%  & 90.88 & 87.71 & 89.08 & 63.67 & 94.79 & 48.89 & 84.13 & 38.87\\
& 50\% & 50\%  & 91.56 & 87.03 & 84.08 & 69.25 & 94.60 & 61.23 & 80.40 & 55.00\\
& 50\% & 100\%  & 93.42 & 87.80 & 87.50 & 70.61 & 93.83 & 65.84 & 81.73 & 52.37\\
& 75\% & 50\%  & 91.31 & 86.90 & 83.23 & 68.73 & 95.13 & 63.47 & 79.93 & 48.44\\
& 75\% & 100\%  & 92.80 & 89.21 & 87.89 & 69.83 & 96.10 & 69.11 & 81.24 & 49.78\\
\bottomrule
\end{tabular}
\caption{ \label{tab1}  
The open intent recognition results of ADB+DeepAligned on four datasets. "KIR" and "LR" mean the known intent ratio and labeled ratio respectively. "Known" denotes the accuracy score on known intents, and "Open" denotes the NMI score on open intents.
}
\end{table*}
\subsubsection{Open Intent Detection}
This module shows the results of identified known intent samples and detected open intent samples. For threshold-based methods, it visualizes the distribution of known and open intents with different confidence scores, which may be helpful for selecting suitable probability threshold, as shown in Figure~\ref{figure-4}. 

For geometrical-based methods,  it visualizes the intent representations on the two-dimension plane. Specifically,  t-SNE~\cite{maaten2008visualizing} is applied to the high-dimension features to achieve the dimensionality reduction. Moreover, we show some auxiliary information of each point (e.g., the centre and radius of ADB), as shown in Figure~\ref{figure-5}.

\subsubsection{Open Intent Discovery}
For unsupervised and semi-supervised clustering methods, it shows the geometric positions of  each produced cluster center with corresponding labels. These centers are categorized into the known and open classes, as shown in Figure~\ref{figure-6}. Users can mine the similarity relations of both known and open intents from observation of center distribution.

As the labels of clusters are not applicable in real scenarios, we adopt the  KeyBERT~\footnote{https://github.com/MaartenGr/keyBERT/} toolkit to extract keywords for open intents in the sentence-level and cluster-level. Furthermore, it calculates the confidence scores of the keywords in the cosine similarity space. The top-3 keywords are recommended for each discovered open intent with respective confidence scores, as shown in Figure~\ref{platform}.
\label{key}

\label{pipeline}

\section{Experiments}

We use four intent benchmark datasets mentioned in section~\ref{data} to verify the performance of our TEXTOIR platform. The known intent ratios are varied between 25$\%$, 50$\%$ and 75$\%$. The labeled proportions are varied between 50$\%$ and 100$\%$. To evaluate the fine-grained performance, we calculate the accuracy score (ACC) on known intents and the Normalized Mutual Information (NMI) score on open intents. We use two state-of-the-art methods of open intent detection and discovery (ADB and DeepAligned) as the components of the pipeline framework. The results are shown in  Table~\ref{tab1}.

The pipeline framework successfully connects two modules, and achieves competitive and robust results in different settings. It essentially overcomes the shortcoming of two modules, and  uses the first module to identify known intents, the second module to discover open intents.

\section{Related Work}
\subsection{Open Intent Detection}
Open intent detection has attracted much attention in recent years. It aims to identify known intents while detecting the open intent. The threshold-based methods use an assigned threshold to detect the open intent. For example, 
MSP~\citep{hendrycks17baseline} computes the softmax confidence score of each known class and regards the low-confidence samples as open. OpenMax~\citep{bendale2016towards} uses the Weibull distribution to produce the open class probability.~\citep{Shu2017DOCDO}  replaces the softmax with the sigmoid activation function and fits Gaussian distribution to the outputs for each known class. ODIN~\citep{liang2018enhancing} adopts temperature scaling and input preprocessing technologies to obtain further discriminative probabilities for detecting open intent. The geometrical feature-based methods use the characteristics of intent features to solve this task. For example, DeepUnk~\citep{lin-xu-2019-deep} first uses the margin loss to learn the discriminative features. Then, it adopts a density-based algorithm, LOF~\citep{breunig2000lof} to discover the anomaly data as the unknown intent. ADB~\citep{Zhang_Xu_Lin_2021} learns the adaptive decision boundary for each known class among Euclidean space. However, all these methods mentioned above fail to discover fine-grained open classes.

\subsection{Open Intent Discovery}
Open intent discovery leverages clustering methods to help find fine-grained clusters as open intents. Unsupervised clustering methods include traditional partition-based method K-Means~\citep{macqueen1967some},  hierarchical method Agglomerative Clustering~\citep{gowda1978agglomerative}, and density-based method~\citep{ester1996density}. There are also clustering methods based on deep neural networks, such as Deep Embedded Clustering (DEC)~\citep{xie2016unsupervised}, joint unsupervised learning (JULE)~\citep{yang2016joint}, and Deep Clustering Network (DCN)~\citep{yang2017towards}. 

As unsupervised methods may not work well on open settings~\citep{lin2020discovering}, researchers try to leverage some prior knowledge to improve the performance. Some methods use pairwise constraints to guide the clustering process, such as KCL~\citep{hsu2018learning}, MCL~\citep{hsu2018multiclass} and CDAC+~\citep{lin2020discovering}. DTC~\citep{Han2019learning} extends DEC with temporal and ensemble information. DeepAligned~\citep{Zhang_Xu_Lin_Lyu_2021}  leverages clustering information to obtain aligned targets for self-supervised feature learning. However, all these clustering methods fail to identify the specific known intent classes.

\section{Conclusion}
We propose the first open intent recognition platform TEXTOIR, which integrates two complete modules: open intent detection and open intent discovery.  It provides toolkits for each module with common interfaces and integrates multiple advanced models and benchmark datasets for the convenience of further research. Additionally, it realizes a pipeline framework to combine the advantages of two modules. The overall framework achieves both identifying known intents and discovering open intents. A series of visualized surfaces help users to manage, train, evaluate, and analyze the performance of different methods.

\section*{Acknowledgments}
This work is founded by National Key R\&D Program Projects of China (Grant No: 2018YFC1707605). This work is also supported by seed fund of Tsinghua University (Department of Computer Science and Technology)-Siemens Ltd., China Joint Research Center for Industrial Intelligence and Internet of Things. We would like to thank the help from Xin Wang and Huisheng Mao, and constructive feedback from Ting-En Lin on this work.

\bibliographystyle{acl_natbib}
\bibliography{anthology,acl2021}

\appendix

\end{document}